\documentclass[11pt,a4paper]{article}
\usepackage[hyperref]{emnlp-ijcnlp-2019}
\usepackage{times}
\usepackage{latexsym}
\usepackage{booktabs}
\usepackage{xspace}
\usepackage{hyperref}
\usepackage{graphicx}
\usepackage{url}
\usepackage{paralist}
\usepackage{multirow}
\usepackage{relsize}
\usepackage[normalem]{ulem}
\usepackage[T1]{fontenc}
\usepackage{amsmath}
\usepackage{amsthm}
\usepackage{amssymb}
\usepackage{expex}
\usepackage{enumitem}
\usepackage{tikz}
\usepackage{amsfonts}
\usepackage{pgfplots}

\DeclareMathOperator*{\E}{\mathbb{E}}
\aclfinalcopy
\usepackage{tikz}
\usetikzlibrary{shapes}
\usetikzlibrary{plotmarks}

\newcommand{\rile}{{RILE}\xspace}
\newcommand{\logic}[1]{\ensuremath{\mathtt{#1}}\xspace}
\newcommand{\Manifesto}{\logic{Manifesto}}
\newcommand{\Party}{\logic{Party}}
\newcommand{\Coalition}{\logic{Coalition}}

\newcommand{\lex}[1]{\textit{#1}\xspace}
\newcommand{\specclass}[1]{\textrm{#1}\xspace}

\newcommand{\tabref}[2][]{Table#1~\ref{#2}\xspace}
\newcommand{\figref}[2][]{Figure#1~\ref{#2}\xspace}
\newcommand{\secref}[2][]{Section#1~\ref{#2}\xspace}

\title{Deep Ordinal Regression for Pledge Specificity Prediction}

\author{Shivashankar Subramanian \qquad Trevor Cohn \qquad Timothy
  Baldwin \\School of Computing and Information Systems\\ The University
  of Melbourne \\
 {\small \url{shivashankar@student.unimelb.edu.au} \qquad \url{{t.cohn, tbaldwin}@unimelb.edu.au}}}

\date{}

\begin{document}
\maketitle
\begin{abstract}

Many pledges are made in the course of an election campaign, forming important corpora for political analysis of campaign strategy and governmental accountability.  At present, there are no publicly available annotated datasets of pledges, and most political analyses rely on manual analysis. In this paper we collate a novel dataset of manifestos from eleven Australian federal election cycles, with over 12,000 sentences annotated with specificity (e.g., rhetorical vs.\ detailed pledge) on a fine-grained scale. We propose deep ordinal regression approaches for specificity prediction, under both supervised and semi-supervised settings, and provide empirical results demonstrating the effectiveness of the proposed techniques over several baseline approaches. We analyze the utility of pledge specificity modeling across a spectrum of policy issues in performing ideology prediction, and further provide qualitative analysis in terms of capturing party-specific issue salience across election cycles.
\end{abstract}

\section{Introduction}
\label{sec:intro}

Election manifestos play a critical role in structuring political campaigns. Campaign communication can influence a party's reputation, credibility, and competence, which are primary factors in voter decision making \cite{fernandez2014and}. Among the various campaign-related functions fulfilled by manifestos \cite{eder2017manifesto}, perhaps the most important is the contract they represent between parties and voters in terms of pledges and prioritisation of political issues \cite{royed2019making}. Political scientists have long studied how specific pledges translate into government programs and actual policy \cite{royed1996testing, thomson2001programme, Naurin2011, schermann2014coalition}. Other work relates specific pledges to the issue clarity of a political party through selective emphasis, which complements salience theory \cite{robertson1976theory, budge1983party, praprotnik2017issue}.
For example:
\begin{quote} 
\lex{we commit ... 30 per cent tax rebate or cash benefit on the cost of private health insurance premiums}
\end{quote}
conveys the party's support for private health insurance, and is more verifiable than:
\begin{quote}
\lex{we will improve the health system.}
\end{quote}
Issue clarity has also been shown to be influenced by a party's ideological position and its role in government \cite{praprotnik2017issue}.

Although pledge specificity prediction is an important task for the
analysis of party position, priorities, and post-election policy
framing, to date, almost all research has relied on manual analysis.
\newcite{SEM2019} is a recent exception to this, in performing speech
act classification over political campaign text, where the class schema
includes the distinction between specific and vague pledges (binary
specificity class).

In this paper, we perform fine-grained pledge specificity prediction, which is more expressive than binary levels \cite{li2016improving, gao2019predicting}.
We use a class schema proposed by \citet{Pomper1980} as detailed in \tabref{tab:speceg}, which captures seven levels of specificity, forming a non-linear increasing order of commitment and specificity \cite{Pomper1980}. Given the non-linear nature of the scale, we use deep ordinal regression models for this task, with distributional loss \cite{imani2018improving}, where we model the output as a uni-modal distribution \cite{beckham2017unimodal}. Our goal is to capture the intuition that a pledge with specificity level $k$, has higher commitment than all the levels $< k$, producing a smoothly varying prediction over the ordinal classes. This can be modeled as a uni-modal distribution which has a probability mass that decreases on both sides of the most probable class. Lastly, as it is expensive to obtain large-scale annotations, in addition to developing a novel annotated dataset, we also experiment with a semi-supervised approach by using unlabeled text.

The contributions of this paper are as follows: (1) we develop and release a dataset\footnote{\url{https://github.com/shivashankarrs/Pledge-Specificity}} for fine-grained pledge specificity prediction based on election manifestos covering eleven Australian federal election cycles (1980--2016), from the two major political parties --- Labor and Liberal; (2) we propose to use deep ordinal regression models for the prediction task, and
evaluate the model under sparse supervision scenarios using the teacher--student framework; and (3) we evaluate the utility of pledge specificity towards ideology prediction, and provide further qualitative analysis by correlating model predictions with party-specific issue salience across major policy areas.

\begin{table*}[!t]
\centering
\begin{small}
\begin{tabular}{l p{5cm} p{6cm} c}
\toprule
 Category & Definition & Example & $\#$\\
\midrule
\specclass{Not a pledge} & Provide facts; greetings; approval or criticism of policies & \lex{We have modernised Australia's industrial relations system, particularly through the 1996 Workplace Relations Act}
& 1  \\ \\[-0.94em]
\specclass{Rhetorical pledge} & Based on moral values and applies to all irrespective of the party  & \lex{We will put our country first} & 2 \\ \\[-0.94em]
\specclass{General pledge} & Specify intangible goals, and also not the ways to achieve them & \lex{Labor will build a stronger and more productive economy} & 3  \\ \\[-0.94em]
\specclass{Continuity pledge} & Commit to the maintenance of currently functioning policy & \lex{We will retain the voluntary health insurance system which now covers the great majority of Australians}
& 4 \\ \\[-0.94em]
\specclass{Goal pledge} & Provide tangible outcomes and goals, without providing the means to achieve them & \lex{A Shorten Labor Government will create 2000 jobs in Adelaide} &  5 \\ \\[-0.94em] 
\specclass{Action pledge} & Provide means to achieve the objective, but don't reveal specific details & \lex{We pledge to support effective voluntary family planning, and to recognize officially the link between social and economic development and the willingness of the individual to limit family size} & 6 \\ \\[-0.94em]
\specclass{Detailed pledge} & Provide clear details of action to achieve an objective & \lex{A re-elected Coalition Government will invest \$1 million to support the Exeter Community Precinct}
& 7  \\ \\[-0.94em]
\bottomrule
\end{tabular}
\end{small}
\caption{Pledge specificity category, definition, example manifesto sentence, and the corresponding specificity value ($\#$).}
\label{tab:speceg}
\end{table*}

\section{Related Work}

Political manifesto text analysis is a relatively novel application, at the intersection of Political Science and NLP. Research has focused primarily on fine-grained policy topic classification and overall ideology prediction tasks \cite{CMP, verberne2014automatic, zirn2016classifying, subramanian2018hierarchical}. Most work dealing with pledge specificity analysis in manifestos has been based on manual analysis, as outlined in \secref{sec:intro}.

Specificity is a pragmatic property of text which has been studied across various fields of research. In
cognitive linguistics, \citet{dixon1987processing} showed that specificity of information in text impacts reading comprehension speed. In Political Science, it has been used to analyze salience, party position and post-election policy framing (see \secref{sec:intro}). There has also been research on the association between text specificity and communication style. In terms of automated specificity analysis, \citet{cook2016content} found specificity in the context of congressional hearings to vary between speakers belonging to the same vs.\ different ideologies. Namely, it was shown that specificity increases as the ideological distance between the committee chair and the witness decreases. 
\citet{SEM2019} addressed two levels of pledge specificity, as part of speech act classification task. Specificity has also been studied in news \cite{louis2011automatic} and classroom discussion domains \cite{luo2016determining, lugini2017predicting}.

These studies have dealt with a restrictive coarse-level analysis (2--3 categories), whereas a fine-grained scale better captures and allows for comparison of election manifestos \cite{Pomper1980}. \newcite{gao2019predicting} was the first attempt at fine-grained text specificity prediction, in the context of social media posts. Here, we target the novel task of fine-grained pledge specificity prediction, which can be used in a range of downstream applications, including capturing party priorities (salience) and ideological position across election cycles. 

All the text specificity analysis work in NLP has modeled the task as classification or regression. As the 7-step pledge specificity levels used in this research \cite{Pomper1980} do not form a single real-valued scale, we model it as an ordinal regression task. Some examples of ordinal regression tasks include sentiment rating prediction \cite{rosenthal2017semeval}, stages of disease prediction \cite{gentry2015penalized}, and age prediction \cite{eidinger2014age}. Recent work has shown that adding a distributional (auxiliary) loss alongside a regression loss, and using expectation to obtain the predicted value \cite{imani2018improving}, provides label smoothing and improves regression performance \cite{gao2017deep}. Approaches based on a uni-modal probability distribution (e.g., Poisson) as output \cite{da2008unimodal, beckham2017unimodal} can be seen as related to the former approach \cite{imani2018improving} where the discrete probability mass function replaces the histogram density. We propose to use a uni-modal distributional loss-based ordinal regression for pledge specificity prediction. 

Secondly, as it is difficult to obtain large amounts of labeled data, existing approaches have used semi-supervised learning \cite{li2015fast, SEM2019}. Here we use a cross-view training approach \cite{clark2018semi, SEM2019}, where we enforce consensus between the intermediate class distributions or the final real-valued output.

\section{Pledge Specificity Dataset}

\begin{table}[!t]
\centering
\begin{small}
\begin{tabular}{c r r c}
\toprule
Specificity & \# Sentences & \% & Avg. Length\\
\midrule
1 & 8165 & 67.00 & 19.5 \\ \\[-0.9em]
2 & 423 & 3.47 & 22.0 \\ \\[-0.9em]
3 & 950 & 7.80 & 23.4  \\ \\[-0.9em]
4 & 300 & 2.46 & 24.6  \\ \\[-0.9em]
5 & 473 & 3.88 & 24.8  \\ \\[-0.9em]
6 & 901 & 7.39 & 25.2 \\ \\[-0.9em]
7 & 973 & 7.99 & 28.5 \\ \\[-0.9em]
\bottomrule
\end{tabular}
\end{small}
\caption{Distribution and length statistics across specificity categories.}
\label{tab:specdis}
\end{table}

\begin{figure}
  \begin{center}
    \includegraphics[scale=0.66]{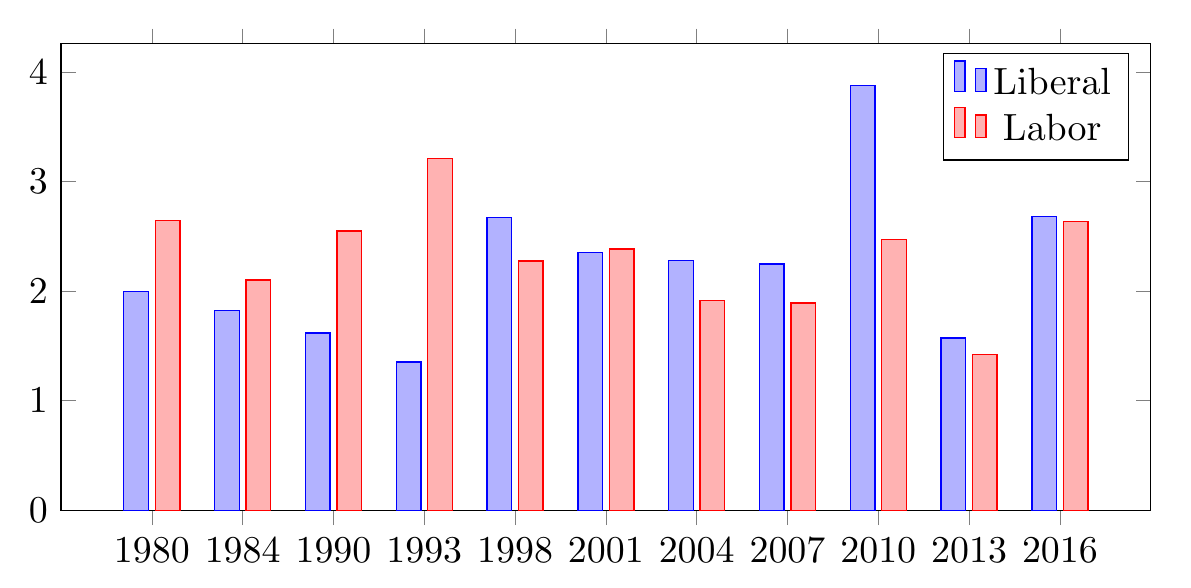}
  \end{center}
\caption{Average pledge specificity of Labor and Liberal parties over different election cycles}
\label{fig:dtp}
\end{figure}

We annotated 22 election manifestos from the Australian Labor and Liberal parties, covering eleven Australian federal election cycles from 1980--2016. The dataset has 12,185 sentences annotated with seven levels of specificity \cite{Pomper1980}. 
See \tabref{tab:speceg} for class definitions and an example of each class. We obtained annotations using the Figure Eight crowdsourcing platform. For each sentence we provided the previous two sentences from the manifesto (as context), the party which published the manifesto, election year, and incumbent and opposition party details. Each sentence was annotated by at least 3 workers after passing quality control (at least 70\% accuracy on test questions). After obtaining annotations, the label which has the highest confidence score is chosen for each sentence. Confidence is the level of agreement between multiple contributors, weighted by the contributors' trust scores. Overall agreement based on the Krippendorf's Alpha is $\alpha=0.58$, indicating moderate agreement \cite{artstein2008inter, krippendorff2011computing}, on par with related studies \cite{gao2019predicting}. The class distribution in the final dataset is given in \tabref{tab:specdis}, alongside the average sentence length in tokens. It can be seen that more specific pledge categories have higher average length. Average specificity values of Labor and Liberal party manifestos across elections are given in \figref{fig:dtp}. The length of the manifesto (in terms of number of sentences) influences average specificity values, with exceptions such as the Liberal party's 2010 election manifesto which is the shortest document but has the highest average pledge specificity value. Further detailed analysis, to decipher the pledge specificity trends in general, is a potential task for future work.

\section{Proposed Approach}
\label{sec:pa}


\subsection{Base Model}

We first obtain representations for each sentence via a sequence of word embeddings, to which we apply a bidirectional GRU (``biGRU'': \citet{cho2014learning}), and concatenate the final hidden state of both the forward and backward GRUs, \mbox{$\mathbf{h}_{i} = \left[\overrightarrow{\mathbf{h}}_{i}, \overleftarrow{\mathbf{h}}_{i}\right]$}. Rather than using a linear activation layer for the output, we study the effect of learning a distribution over ordinal classes, and using an expectation layer to get the final prediction, which we now expound upon.

\subsection{Distributional Loss}
\label{sec:dl}

\begin{figure}[!t]
  \begin{center}
    \includegraphics[scale=0.38]{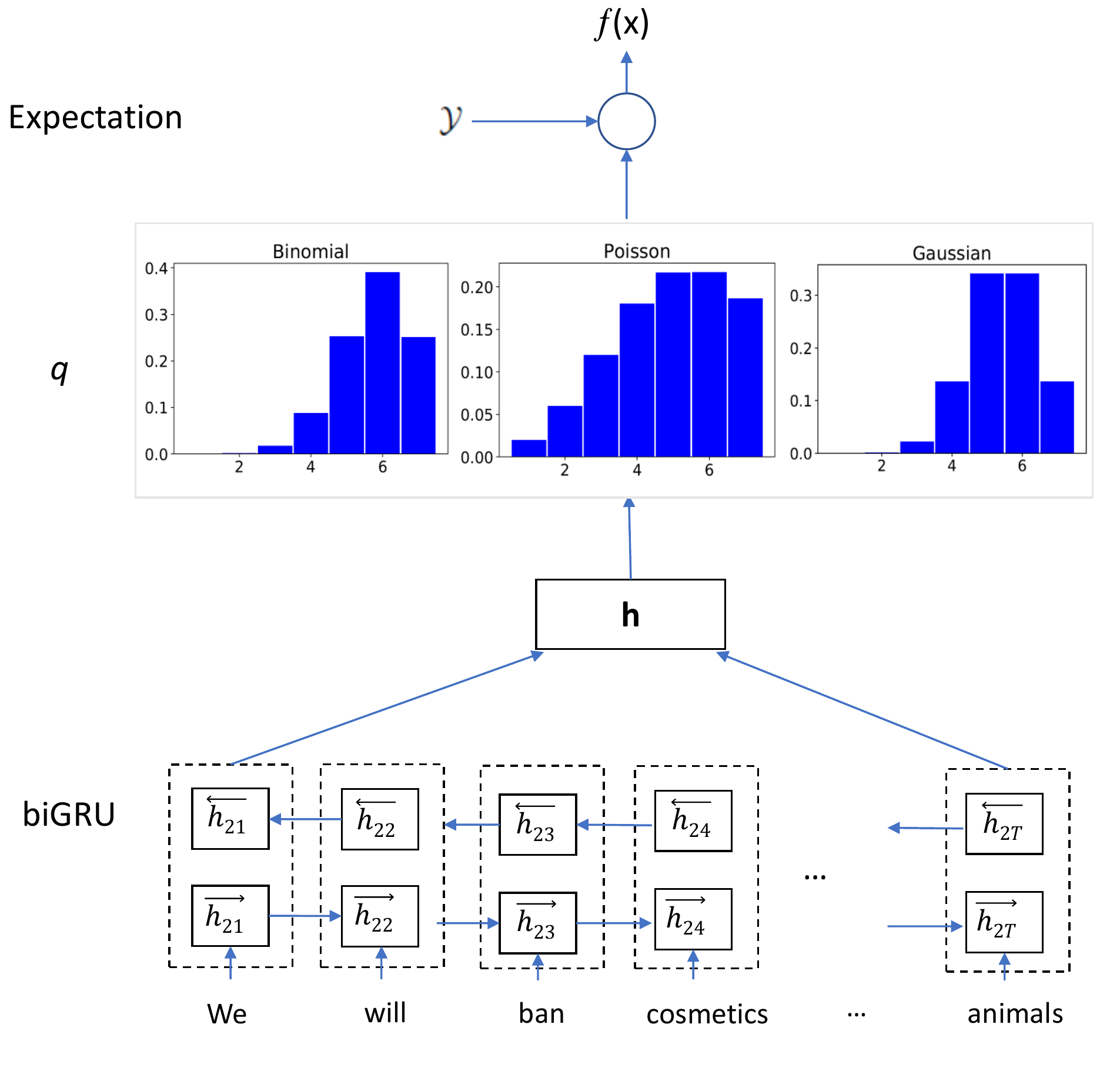}
\end{center}
  \caption{Illustration of the model architecture, comprising a biGRU over sentence tokens, to compute the parameter of one of the three distributions: $p$ for Binomial (with $n =\ K-1$) and $\lambda$ for Poisson. \emph{Pmf} of these distributions are then used to define a categorical distribution over the ordinal classes $\mathcal{Y}$. For learning, we use the categorical cross-entropy against the gold \emph{one-hot} $y$, as well as the squared error $\left(y - f(x)\right)^2$, where $f(x)$ is given by expectation taken under the predicted categorical distribution $q$. Gaussian uses a different mechanism, as described in \secref{gauss} where the categorical distribution ($q$) is predicted directly using a $K$-dimensional softmax output, and the cross-entropy is computed between $q$ and a Gaussian histogram density centred at $\mu=y$, discretised by way of integration of the PDF between adjacent label indices.}

\label{fig:OverallModel}
\end{figure}

Let us assume that the continuous target variable $Y$ is normally distributed, conditioned on inputs $x \in \mathbb{R}^d$, $Y\! \sim\! \mathcal{N}(\mu\!=\!f(x), \sigma^2)$ for a fixed variance $\sigma^2$ > 0. In regression, the maximum likelihood function $f$ for $n$ samples \{$x_i$, $y_i$\} corresponds to minimizing $l_2$ loss, such that $f(x) = \E(Y|x)$. Alternatively, we can learn a categorical distribution ($q_{x}$) over the ordinal classes $\mathcal{Y}$, and use the expected value as the prediction, $f(x)$ \cite{rothe2018deep}. In this work, we follow the latter method, but parameterise the categorical distribution based on uni-modal probability distribution, a technique which has been shown to perform well for ordinal regression tasks \cite{beckham2017unimodal}. This modification converts the problem to a more difficult (multi-task) problem, that promotes generalization and reduces over-fitting \cite{imani2018improving}. The overall objective is to jointly minimize the squared loss for the regression task ($\mathcal{L}_{S}$), and cross-entropy for the distributional loss over $\mathcal{Y}$ ($\mathcal{L}_D$), based on the objective $\mathcal{L}_J = \alpha \mathcal{L}_{S} + \mathcal{L}_D$, where the hyper-parameter $\alpha$ is tuned using a validation set. We experiment with different distributions in generating the intermediate representations $q_{x}$, including categorical (as a baseline approach, see Section \ref{sec:sup}: \newcite{beckham2016simple, gao2017deep, rothe2018deep}), discrete uni-modal (Binomial and Poisson: \newcite{beckham2017unimodal}), and truncated Gaussian \cite{imani2018improving}. The final prediction is obtained using expectation, which has been shown to be effective for various regression tasks in the vision domain. Here we study the use of uni-modal distributional loss-based ordinal regression approaches \cite{beckham2017unimodal, imani2018improving} for text specificity analysis (Section \ref{sec:sup} has results demonstrating its superiority over the other choices). We detail the different ways to obtain $q_{x}$, and the corresponding loss functions $\mathcal{L}_D$ below, and provide an overall summary in \figref{fig:OverallModel}.

\subsubsection{\textsc{Binomial}}
\label{binomial}
\label{sec:bin}

With the biGRU model, we estimate the parameter ($p$) of the Binomial distribution (with a sigmoid output), based on which the distribution over classes can be obtained via the probability mass function,
\[p(k; K - 1, p) = \binom{K-1}{k} p^k (1 - p)^{K-1-k}, \]
$k \in \{0, 1, \ldots K-1\}$, where $p$ $\in$ [0, 1]. As the final layers (post sigmoid) are under-parametrized, we have a softmax layer with $\tau$ after obtaining the probability masses, 
\[q_k = \frac{\exp(\phi_k/\tau)}{\sum^{K-1}_{j=0} \exp(\phi_j/\tau)}, \]
where  $\tau \sim \text{SoftPlus}(\tau')$, and $\tau'$ is learned by the deep net, conditioned on the input ($x$). We then have an expectation layer to obtain the final output $f(x)$. 
Output of the softmax layer is fit to the one-hot encoded ordinal classes for each input ($\textbf{y}$), by minimizing the cross-entropy loss ($\mathcal{L}_{\mathcal{D}\textsc{Binomial}}$). 

\subsubsection{\textsc{Poisson}}
\label{poisson}
\textsc{Poisson} is similar to the binomial case, in that we obtain the parameter ($\lambda$) of the Poisson distribution using the biGRU, with a $\text{SoftPlus}$ activation. We then use the probability mass function of the Poisson distribution to get the probabilities over different classes, $k \in \mathcal{Y}$,
\[ 
p(k; \lambda) = \frac{\lambda ^ {\kern 0.08 em k} e ^ {-\lambda}} {k!}, \]
which is again passed through a softmax layer to obtain $q_{x}$, fit by minimizing cross-entropy loss ($\mathcal{L}_{\mathcal{D}\textsc{Poisson}}$), and an expectation layer is used to obtain the final prediction. 

\subsubsection{Gaussian (\textsc{Gauss})}
\label{gauss}
To compute $\E(Y|x)$ ($\mu$ of the Gaussian), 
here we fit the intermediate distribution $q_{x}$ directly to histogram density of a truncated Gaussian distribution with support $[1, K]$ (target distribution: $p_{*}$). We achieve this by learning a prediction distribution with the biGRU model, 
$q_{x}$ : $\mathcal{Y}$ $\rightarrow [0,1]$. For this, the  ordinal label of training instances is transformed into a truncated Gaussian PDF. The mean $\mu$ for this Gaussian is the target $y$ of each data-point, with fixed variance $\sigma^2$, which we set to the radius of the bins in $\mathcal{Y}$ (1 in this case). 
The CDF for the chosen target distribution is computed as $\frac{1}{2} (1+\text{erf}(\frac{x-\mu}{\sigma\sqrt{2}}))$ and $p_{*}$ is obtained for each class, $\text{k} \in \mathcal{Y}$, as,
\[\frac{1}{2} \bigg(\text{erf}\big(\frac{k-\mu}{\sigma\sqrt{2}}\big)-\text{erf}\big(\frac{k-1-\mu}{\sigma\sqrt{2}}\big)\bigg).\]
This formulation allows efficient computation of divergence between $p_{*}$ and $q_{x}$ for optimization, which results in cross-entropy minimization ($\mathcal{L}_{\mathcal{D}\textsc{Gauss}}$: \newcite{imani2018improving}). Note that the training target $p_{*}$ is uni-modal, and no constraints are \textit{explicitly} enforced on the shape of $q_{x}$. 


\subsection{Incorporating Context}

We incorporate context in the form of information from adjacent sentences following the approach of \newcite{liu2017using}: for each training sentence, we use the predicted (intermediate) probability distribution across ordinal classes of the previous $L$ sentences as context. A new biGRU model is trained with the sentence and the additional contextual information, concatenated to $\mathbf{h}_{i}$. We refer to this model as $\text{biGRU}_{\textsc{ord\! \text{+}\! context}}$. In the test phase, biGRU$_{\textsc{ord}}$ provides contextual information, and the newly trained model ($\text{biGRU}_{\textsc{ord\!  \text{+}\!  context}}$) is used to predict the test sentence output.

\subsection{Semi-supervised Learning}
\label{ssl}
As it is expensive to get large-scale specificity annotations we employ a cross-view training approach \cite{clark2018semi, SEM2019} for semi-supervised learning, which can leverage additional unlabeled text. Cross-view training is a kind of teacher--student method, whereby the model ``teaches'' a ``student'' model to classify unlabelled data. The student has a restricted view over the data, e.g., through the application of noise \cite{sajjadi2016regularization, wei2018improving}. 
We use biGRU$_{\textsc{ord\! \text{+}\! context}}$ with word-level dropout and zero vector set to contextual information as the auxiliary model. This procedure regularizes the learning of the teacher to be more robust, as well as increasing its exposure to unlabeled text. 
We augment our dataset with over 32k sentences from UK and US election manifestos released from the same time period. On these unlabeled examples, the model's output is used to fit the auxiliary model by enforcing consensus in their predictions. This consensus loss $\mathcal{L}_{\text{U}}$ is added to the supervised training objective ($\mathcal{L}_{J}$). Under the semi-supervised setting, we evaluate the following  approaches: 
\begin{description}[nolistsep]
\item{\textsc{Mse:}} use the final regression output of the teacher model ($f(x)$) to fit an auxiliary model, thereby enforcing consensus using a squared loss, $\text{MSE}(\E_{q_{\theta}}[\mathcal{Y}], \E_{q_{\omega}}[\mathcal{Y}])$ where $\mathcal{Y}$ is a fixed class vector; denoted as ``$\mathcal{L}_{\text{UMSE}}$''. 
\item{\textsc{Kld:}} an intermediate distribution over targets $q_{\theta}(\mathcal{Y}|s)$ is used to fit an auxiliary model, $q_{\omega}(\mathcal{Y}|s)$, by minimising the Kullback-Leibler (KL) divergence, $\text{KL}(q_{\theta}(\mathcal{Y}|s), q_{\omega}(\mathcal{Y}|s))$; denoted as ``$\mathcal{L}_{\text{UKLD}}$''.\footnote{This is the closest setting to \newcite{SEM2019}, which minimizes KL divergence between output distributions in a classification setting. But the overall objective is different in our case, in that we have an expectation layer over $q$ to obtain the target regression output.} 
\item{\textsc{Emd:}} $q_{\theta}(\mathcal{Y}|s)$ is again used to fit the auxiliary model, $q_{\omega}(\mathcal{Y}|s)$, by minimising the earth mover's distance, $\text{EMD}(q_{\theta}(\mathcal{Y}|s)$, $q_{\omega}(\mathcal{Y}|s))$; denoted as ``$\mathcal{L}_{\text{UEMD}}$''. EMD is defined as $ \text{EMD}(q_{\theta}, q_{\omega}) = \frac{1}{K}^{\frac{1}{l}} \|\text{cmf}(q_{\theta}) - \text{cmf}(q_{\omega})\|_{l} $,
where $\text{cmf}(\cdot)$ is the cumulative mass function for the predicted (intermediate) probability distribution $q$, and we use $l=2$. EMD considers distance between classes, and is more suitable for ordinal tasks \cite{hou2016squared}. 
\end{description}

\section{Experimental Results}
\label{sec:sup}
To evaluate model performance we use macro-averaged mean absolute error (MMAE: \newcite{rosenthal2017semeval}) given the class imbalance, and Spearman's $\rho$. MMAE is given as $\frac{1}{|K|}\! \sum \limits_{j=1}^{|K|} \frac{1}{|S_{j}|}\! \sum \limits_{x_{i} \in S_{j}} |f(x_{i}) - y_{i}|$, 
where $S_{j}$ denotes the subset of instances annotated with (true) ordinal class $j$. We consider the following baselines:
\begin{description} [nolistsep]
\item {\textbf{Majority}}: assign the majority class in the training set to all test instances.
\item {\textbf{Length}}: use sentence length as the specificity score.
\item \textbf{Speciteller}: co-training model of \newcite{li2015fast}, used by \citet{cook2016content} for congressional hearings specificity analysis. 
\item \textbf{NN$_{\textsc{reg}}$}: bag-of-words term-frequency representation,  fed into a feed-forward neural network model \cite{gao2019predicting}.
\item \textbf{biGRU$_{\textsc{reg}}$}: biGRU model trained with a mean squared loss objective.
\item \textbf{biGRU$_{\textsc{class}}$}: biGRU model trained with a cross-entropy objective \cite{SEM2019}.
\item \textbf{biGRU$_{\textsc{reg}_{l_1}}$}: biGRU regression model with mean absolute error objective ($l_1$ loss). 
\end{description}

\begin{table}[!t]
      \centering
      \begin{smaller}
      \scalebox{1}{
     \begin{tabular}{l l c c}
    \toprule
       Category & Approach & MMAE & $\rho$ \\
       \midrule  
       & Majority  & 3 & - \\
       & Length  & -  &  0.21  \\
       & Speciteller & - &  0.18 \\     
       & NN$_{\textsc{reg}}$ & 2.05 &  0.33   \\
       & biGRU$_{\textsc{reg}}$  & 1.83 &  0.47   \\
       & biGRU$_{\textsc{class}}$  & 2.17 & 0.40 \\ 
       & biGRU$_{\textsc{reg}_{l_1}}$  & 1.99 & 0.46 \\
       \midrule
      Ordinal & biGRU$_{\textsc{Categorical}}$  & 1.80 &0.48  \\
      Ordinal   & biGRU$_{\textsc{Binomial}}$  &1.78 &0.48  \\
    Ordinal   & biGRU$_{\textsc{Poisson}}$  & 1.90 &0.41  \\
    Ordinal   & biGRU$_{\textsc{Gauss}}$  & 1.72 & \textbf{0.49}  \\
    \midrule
    Ordinal   & biGRU$_{\textsc{Gauss + context}}$  &\textbf{1.71} &\textbf{0.49}  \\
     \bottomrule
   \end{tabular}
   }
  \end{smaller}
  \caption{Specificity prediction performance; the best approach is given in bold.}
  \label{tab:res}
\end{table}

\begin{figure*}[!t]
\centering
 \begin{minipage}{.5\textwidth}
  \centering
  \includegraphics[scale=0.7]{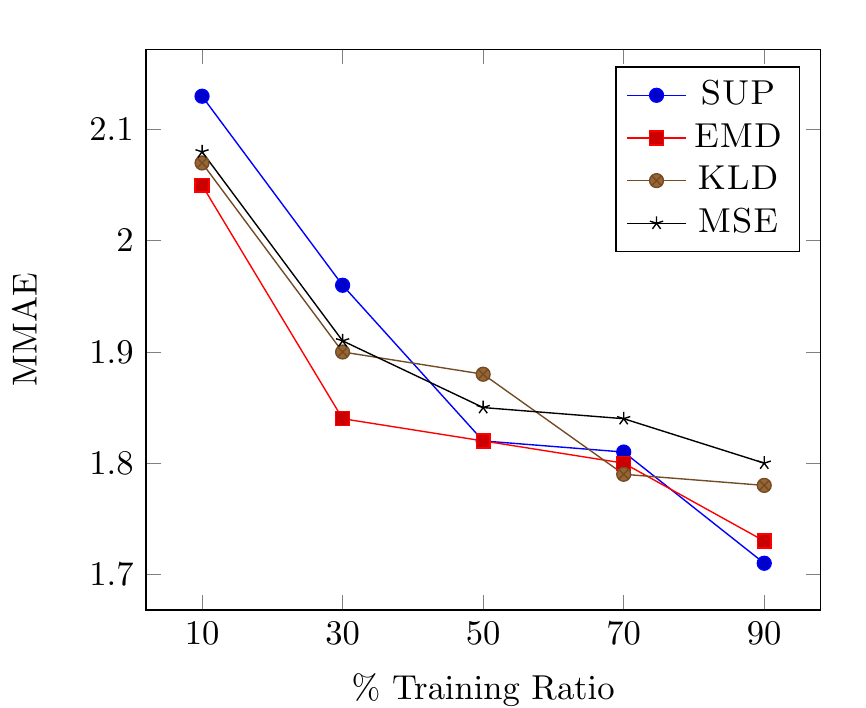}
\end{minipage}%
\begin{minipage}{.5\textwidth}
  \centering
  \includegraphics[scale=0.7]{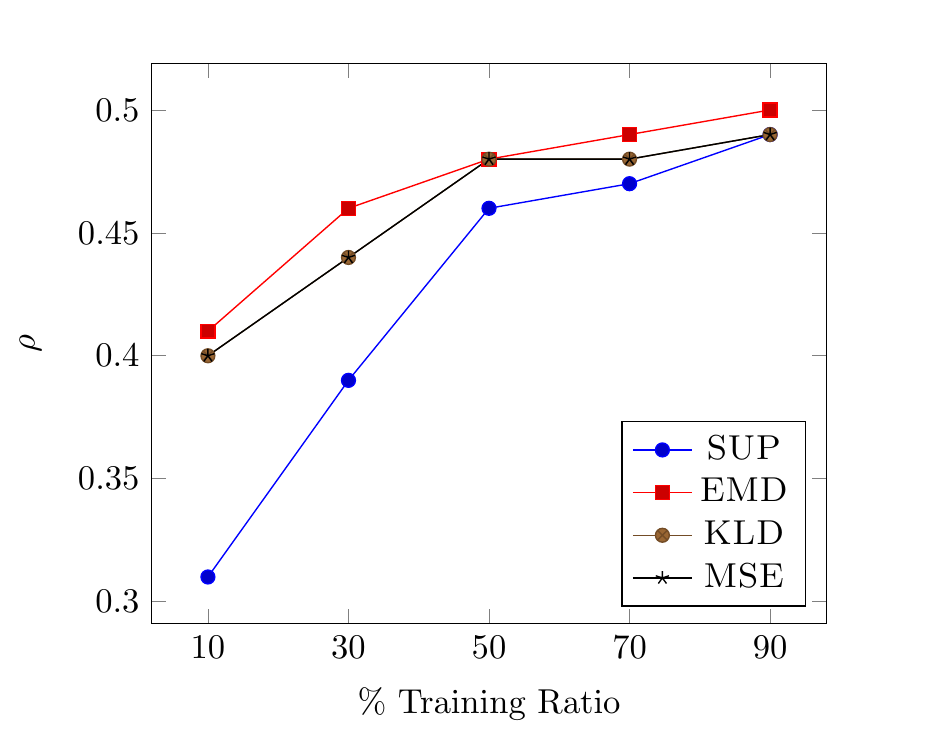}
  \end{minipage}
   \begin{minipage}{.5\textwidth}
     \end{minipage}
\captionof{figure}{Prediction performance across different training ratios. Note that 90\% =  all the training data, as 10\% is used for validation. The supervised ordinal model (\textsc{Sup}) and semi-supervised teacher--student models (\textsc{Mse}, \textsc{Kld}, \textsc{Emd}, given in \secref{ssl}) are compared on MMAE and Spearman's $\rho$.}
\label{ssl:results}
\end{figure*}

All the baseline and proposed biGRU models use ELMo embeddings  \cite{peters2018deep}. The regression models minimize $l_2$ loss, unless otherwise specified.  We compare the average performance across five runs with an 80:20 train:test split. We randomly choose 10\% of instances from the training set as validation data. We compare the baseline approaches with our proposed ordinal approaches, which have an intermediate distributional loss in conjunction with the final prediction loss: biGRU$_{\textsc{Gauss}}$ (\secref{gauss}),  biGRU$_{\textsc{Binomial}}$ (\secref{binomial}), or biGRU$_{\textsc{Poisson}}$ (\secref{poisson}). We also evaluate biGRU$_{\textsc{Categorical}}$, where the softmax layer is fitted to one-hot encoded class labels \cite{gao2017deep, rothe2018deep}. Note that this is not uni-modal.

\newcite{gao2019predicting} used a combination of bag-of-words representation, surface features, social-media-specific features (eg., Tweet mentions), and emotion-related features, with a support vector regression model which minimizes squared loss. Social media and emotion-related attributes are not relevant to our data, and other surface features did not provide improvements. Hence we show the performance of the bag-of-words representation with squared loss objective (NN$_{\textsc{reg}}$ in Table \ref{tab:res}). From the results in \tabref{tab:res}, we can see that sequential models with ELMo embeddings (biGRU) perform better than neural bag-of-words models (NN$_{\textsc{reg}}$). The $l_{2}$ regression model (biGRU$_{\textsc{reg}}$) performs better than $l_{1}$ regression (biGRU$_{\textsc{reg}_{l_1}}$) and classification (biGRU$_{\textsc{class}}$). 

With respect to deep ordinal approaches, biGRU$_{\textsc{Poisson}}$ performs better than classification (biGRU$_{\textsc{class}}$), but does not improve upon regression (biGRU$_{\textsc{reg}}$). The Binomial performs better than the Poisson, consistent with previous work \cite{beckham2017unimodal, da2008unimodal}. biGRU$_{\textsc{Categorical}}$ performs better than biGRU$_{\textsc{reg}}$, but not over unimodal approaches (biGRU$_{\textsc{Binomial}}$, biGRU$_{\textsc{Gauss}}$).
Overall, the model which fits intermediate distribution to a truncated Gaussian (histogram density) target distribution provides the best performance. It gives over 6\% improvement in terms of MMAE, and over 4\% in $\rho$, compared to biGRU$_{\textsc{reg}}$. Adding context to biGRU$_{\textsc{Gauss}}$ (biGRU$_{\textsc{Gauss\! +\! context}}$) provides a slight reduction in error. 

\subsection{Semi-supervised Learning}
We next compare the performance of biGRU$_{\textsc{Gauss\! +\! context}}$ (\textsc{Sup}) and the semi-supervised extensions of it (\secref{ssl}) which leverage additional unlabeled data: minimizing $\mathcal{L}_{\text{UMSE}}$ (\textsc{Mse}), 
$\mathcal{L}_{\text{UKLD}}$
(\textsc{Kld}), and $\mathcal{L}_{\text{UEMD}}$ (\textsc{Emd}). The amount of labeled data in the training split is varied from 10\% to 90\%. Results are presented in \figref{ssl:results} for MMAE and $\rho$. From the results, semi-supervised approaches provide large gains in terms of both MMAE and $\rho$, especially when training with fewer instances, ratio $\leq$ 30\%. Overall, the semi-supervised learning approach which minimizes \textsc{Emd} performs best across all training ratios compared to both supervised and other semi-supervised approaches. It provides .10 and .06 absolute improvements in $\rho$ under sparse supervision scenarios (10\% and 30\% of training data, resp.). Even under richer supervision settings ($\geq$ 70 \%), it provides higher $\rho$.
\section{Political Analysis Using the Models}
\label{ds}

Political scientists utilize pledge specificity for a variety of applications (see \secref{sec:intro}). Here, we extrinsically evaluate our specificity model using two tasks related to campaign strategy: (1) party position or ideology prediction (\secref{ip}), and (2) issue salience analysis (\secref{sec:sal}). For both tasks, we compare the use of pledge specificity across policy issues vs.\ a count-based representation of  policy mentions. 

\subsection{Ideology Prediction}
\label{ip}
Estimating the manifesto-level ideology score on the left--right spectrum using sentence-level policy topic annotations is a popular task \cite{slapin2008scaling, lowe2011scaling, daubler2017estimating, subramanian2018hierarchical},  for which the policy scheme provided by CMP \cite{CMP} is commonly used. It has 57 political themes, across 7 major  categories. Among those approaches, the \rile index is the most widely adopted \cite{merz2016manifesto, oxford}, and has been shown to correlate highly with other popular scores \cite{lowe2011scaling}. \rile is defined as the difference between count of (pre-determined) right and left policy theme mentions across sentences in a manifesto \cite{budge2013}. Here we evaluate the effectiveness of using the proposed specificity modeling across those policy issues, compared to using \rile-based party position scores \cite{budge2013}. 

We compute the specificity weight \cite{Pomper1980} from the average specificity score across sentences, $\frac{1}{|I|} \sum \limits_{S_{i} \in I} \text{Spec}(S_{i})$ for each policy issue ($I$). With specificity weight as the basic feature, we also model global signals such as party coalition and temporal dependencies across elections, which can enforce smoothness in manifesto positions \cite{greene2016competing, subramanian2018hierarchical} based on probabilistic soft logic.



\subsubsection{Probabilistic Soft Logic}

To address this, we propose an approach using hinge-loss Markov random fields (``HL-MRFs''), a scalable class of continuous, conditional graphical models \cite{bach2013}. These models can be specified using Probabilistic Soft Logic (``PSL'': \newcite{bach2015}), a weighted first order logical template language. An example of a PSL rule is $\lambda: \mathtt{P(a)} \wedge \mathtt{Q(a, b)} \rightarrow \mathtt{R(b)}$,
where $\mathtt{P}$, $\mathtt{Q}$, and $\mathtt{R}$ are predicates, $\mathtt{a}$ and $\mathtt{b}$ are variables, and $\lambda$ is the weight associated with the rule. PSL uses soft truth values for predicates in the interval $\big[0,1\big]$. The degree of ground rule satisfaction is determined using the Lukasiewicz $t$-norm and its corresponding co-norm as the relaxation of the logical AND and OR, respectively. The weight of the rule indicates its importance in the HL-MRF probabilistic model, which defines a probability density function of the form:
\[ \!
  P(\mathbf{Y}|\mathbf{X}) \propto \exp \left(-\sum_{r=1}^{M} \lambda_{r} \phi_{r}(\mathbf{Y}, \mathbf{X}) \right), 
\]
where $\phi_r(\mathbf{Y}, \mathbf{X})\!$  $=$   $\!\max{\{l_{r}(\mathbf{Y}, \mathbf{X}), 0 \}}^{\rho_{r}}$ is a hinge-loss potential corresponding to an instantiation of a rule, and is specified by a linear function $l_{r}$ and optional exponent $\rho_r$ $\in \{1, 2\}$. 

\begin{table*}[!t]
 \begin{small}
 \centering
  \scalebox{1}{
  \begin{tabular}{p{15.6cm}}
   \toprule  \textbf{Specificity Weight --- Model I} \\
   \midrule

 $\mathtt{Manifesto(x)} \wedge \mathtt{Policy(I)} \wedge \mathtt{Specw(x, I)} \wedge \mathtt{IdeologyMap(I, social\ left/right)}   \rightarrow \mathtt{socpos(x)}$ \\
 $\mathtt{Manifesto(x)} \wedge \mathtt{Policy(I)} \wedge \mathtt{Specw(x, I)} \wedge \mathtt{IdeologyMap(I, economic\ left/right)}  
 \rightarrow \mathtt{econpos(x)}$ \\
 \midrule
 \textbf{Overall position ---  Model II} \\
   \midrule

    $\mathtt{Manifesto(x)} \wedge \mathtt{socpos(x)} \rightarrow \mathtt{pos(x)}$ \\
    $\mathtt{Manifesto(x)} \wedge \mathtt{econpos(x)}  \rightarrow \mathtt{pos(x)}$ \\
    \midrule
 \textbf{Global signals ---  Model III} \\
   \midrule
   $\mathtt{Manifesto(x)} \wedge \mathtt{Party(x, a)} \wedge \mathtt{Manifesto(y)} \wedge \mathtt{Party(y, b)} \wedge \mathtt{Coalition(a, b)} \wedge \mathtt{pos(x)} \rightarrow \mathtt{pos(y)}$ \\
   $\mathtt{Manifesto(x)} \wedge \mathtt{Party(x, a)} \wedge \mathtt{PreviousManifesto(x, t)} \wedge \mathtt{Party(t, a)} \wedge \mathtt{pos(t)} \rightarrow \mathtt{pos(x)}$ \\    
   \midrule
 \textbf{Relative specificity ---  Model IV} \\
 \midrule
 $\mathtt{Manifesto(x)} \wedge \mathtt{Policy(I)} \wedge \mathtt{SpecScale(x, I)} \wedge \mathtt{IdeologyMap(I, social\ left/right)}   \rightarrow \mathtt{socpos(x)}$ \\
 $\mathtt{Manifesto(x)} \wedge \mathtt{Policy(I)}  \wedge \mathtt{SpecScale(x, I)} \wedge \mathtt{IdeologyMap(I, economic\ left/right)}  
 \rightarrow \mathtt{econpos(x)}$ \\
 \bottomrule
  \end{tabular}
  }
  \caption{PSL Model: Representative rules. \logic{left}/\logic{right} in the \logic{IdeologyMap} predicate indicates policy issues mapped to left/right categories, which is implemented as two separate rules --- one for left and another for right.}
  \label{tab:pslrule}
  \end{small}
\end{table*}

\subsubsection{PSL Model} 
\label{pslmodel}
Here we elaborate on our PSL model based on manifesto content-based features (specificity weight across 57 policy issues), coalition information, and temporal dependencies. Our target \logic{pos} (left--right position) is a continuous variable $\big[0,1\big]$, where 1 indicates an extreme right position, 0 denotes an extreme left position, and 0.5 indicates center. We also model the social and economic positions explicitly (\logic{socpos} and \logic{econpos}), which influence the overall \logic{pos}. Each instance of a manifesto, its party affiliation and policy issues, are denoted by the predicates \Manifesto, \Party and \logic{Policy}. Other predicates are given as follows:

\begin{description}[nolistsep]
\item \textbf{Specificity weight} of each policy issue in the given manifesto (\logic{Specw}).

\item \textbf{Relative specificity scale}: ratio of specificity weight for each policy issue given a party's manifesto, to maximum specificity weight for the same policy issue across parties from the same country and election (\logic{SpecScale}).

\item \textbf{Policy issue mapping}: 26 out of the 57 policy themes are categorized as social and economic left--right issues by \citet{benoit2007estimating} (\logic{IdeologyMap}). 

\item \textbf{Coalition}: captures the strength of ties between two parties, given by a logistic transformation of the number of times two parties have been in a coalition in the past (\Coalition).

\item \textbf{Temporal dependency} between a party's current manifesto position and its previous manifesto position (\logic{PreviousManifesto}).



\end{description}

Representative rules of our PSL model, based on the predicates presented above, are given in \tabref{tab:pslrule}. They include: 
\begin{description}[nolistsep]
\item \textbf{Specificity}: if a manifesto contains more specific pledges related to social (or economic) left/right policies, then it will more likely be a social (or economic) left/right-aligned manifesto.
\item \textbf{Overall position}: social and economic position influences the overall position, and this allows the model to place different weights on the influence of social and economic policies on the overall position, which is found to be necessary by  \citet{benoit2007estimating}. 
\item \textbf{Global signals}: coalition and temporal dependencies to enforce smoothness in manifesto positions. 
\item \textbf{Relative specificity}: \logic{SpecScale} of a left (or right) policy during an election amplifies its overall position scores. 
\end{description}

\begin{table}[!ht]
\centering
\begin{small}
\begin{tabular}{l r r}
\toprule
Model  & Aus & UK \\
\midrule
bootstrapped \logic{pos} & 0.66 & 0.39 \\ \\[-0.9em]
PCA & 0.11 & 0.39 \\ \\[-0.9em]
Model I + II & 0.63 & 0.33  \\ \\[-0.9em]
Model I + II + III & 0.65 & 0.33   \\ \\[-0.9em]
Model I + II + III + IV & \textbf{0.71} & \textbf{0.45}   \\ \\[-0.9em]
\bottomrule
\end{tabular}
\end{small}
\caption{Spearman's $\rho$ for prediction of party position based on the different models.}
\label{tab:psl}
\end{table}

\subsubsection{Evaluation}
We use manifestos from Australia and UK for our analysis. We use data from Voter Survey \cite{Votersurvey} for Australia and CHES Expert Survey \cite{bakker2015measuring} for the UK as the gold-standard party position. A primary step (related to the model given in \secref{pslmodel}) is to obtain policy topic classification for sentences in each manifesto. If annotations are not available from \newcite{CMP}, one out of 57 political themes are predicted using the method of \citet{subramanian2018hierarchical}. Specificity scores of sentences are obtained using the proposed ordinal regression approach (biGRU$_{\textsc{Gauss+context}}$). Using social, economic and a combined list of left--right policy themes (\logic{IdeologyMap}), and with the RILE formulation, we bootstrap \logic{socpos}, \logic{econpos} and \logic{pos}. We then use the PSL model (\tabref{tab:pslrule}) to re-calibrate the scores based on specificity scores and the global signals.

We compare the performance of bootstrapped \logic{pos} (\rile or policy count-based) with the PSL model. Principal component analysis (``PCA'': \citet{PCA}) on the frequency distribution, and projection on its principal component, is used as an additional baseline. Spearman's correlation ($\rho$)  against the gold-standard positions is given in \tabref{tab:psl}. Overall, pledge specificity, especially on a relative scale (which differentiates emphasis between parties) provides large gains, and global signals give only mild improvements. 

\subsection{Capturing Issue Salience}
\label{sec:sal}

For the Australian manifestos (from the Greens, Labor, Liberal, and National parties) we perform a qualitative study of specificity weight across policy themes, by correlating it against the salience of major policy areas given by the Voter Survey \cite{Votersurvey}. Again we compare its utility over the use of counts across policy themes in a manifesto. Using sentences classified with policy themes and specificity scores using our proposed approach, we construct the following $|$Manifestos$|$ $\times$ 57 features --- frequency distribution ($\mathbf{C}$) and pledge specificity weight ($\mathbf{S}$) across policy themes. The features are used as independent variables, and voter survey salience scores across major policy areas --- health, education, environment, tax, and economy --- are treated as dependent variables. Note that the voter survey scores are available for each party and election cycle across policy areas. We build separate multi-variate linear regression models and compare them based on the goodness of fit (log-likelihood). Log-likelihood values are given in \tabref{tab:lr}: across all policy areas, pledge specificity better captures salience than a count-based representation. 

\begin{table}[!t]
\centering
\begin{small}
\begin{tabular}{l c c}
\toprule
 Policy Area & $\mathbf{C}$ & $\mathbf{S}$ \\
\midrule
Health & 824.92 & 905.19 \\ \\[-0.9em]
Education & 781.15 & 905.54  \\ \\[-0.9em]
Environment & 841.89 & 897.53   \\ \\[-0.9em]
Tax & 735.97 & 824.69   \\ \\[-0.9em]
Economy & 258.30 & 276.71   \\ \\[-0.9em]
\bottomrule
\end{tabular}
\end{small}
\caption{Log-likelihood with pledges specificity weight ($\mathbf{S}$) and count of sentences ($\mathbf{C}$) across 57 policy themes as independent variables.  Log-likelihood values using $\mathbf{S}$ are better than $\mathbf{C}$ across all the policy areas.}
\label{tab:lr}
\end{table}

\section{Conclusion and Future Work}
In this work we present a new dataset of election campaign texts, annotated with pledge specificity on a fine-grained scale. We study the use of deep ordinal regression approaches using an auxiliary uni-modal distributional loss for this task. The proposed approaches provide large gains in performance under both supervised and semi-supervised settings. Specificity weight across policy issues benefits ideology prediction and also better captures issue salience, compared to the traditional policy theme count-based representation. This aligns with previous studies done based on manual annotations \cite{praprotnik2017issue}. In future work, we aim to expand this study to multiple languages.

\section*{Acknowledgements}
We thank Robert Thomson for his inputs on the task definition. This work was funded in part by the Australian Government
Research Training Program Scholarship, and
the Australian Research Council. 

\bibliography{emnlp-ijcnlp-2019}
\bibliographystyle{acl_natbib}
\end{document}